%% file: main.tex
\documentclass[10pt,twocolumn,letterpaper]{article}

\usepackage[pagenumbers]{cvpr} 

\makeatletter
\@namedef{ver@everyshi.sty}{}
\makeatother
\usepackage{tikz}

\usepackage{graphicx}
\usepackage{amsmath}
\usepackage{amssymb}
\usepackage{booktabs}
\usepackage{times}
\usepackage[utf8]{inputenc}
\usepackage{algorithm}
\usepackage{algorithmic}
\usepackage{float}
\usepackage{cancel}
\usepackage{multirow}
\usepackage{todonotes}
\usepackage{subcaption}
\usepackage{pgfplots}
\DeclareUnicodeCharacter{2212}{−}
\usepgfplotslibrary{groupplots,dateplot}
\usetikzlibrary{patterns,shapes.arrows}
\pgfplotsset{compat=newest}
\renewcommand{\algorithmiccomment}[1]{\bgroup\hfill//~#1\egroup}
\usepackage{tikz}
\usetikzlibrary{positioning}
\usepackage[pagebackref,breaklinks,colorlinks]{hyperref}

\usepackage[capitalize]{cleveref}
\crefname{section}{Sec.}{Secs.}
\Crefname{section}{Section}{Sections}
\Crefname{table}{Table}{Tables}
\crefname{table}{Tab.}{Tabs.}

\def\cvprPaperID{46} 
\def\confName{CVPR}
\def\confYear{2022}

\begin{document}

\title{$n$-CPS: Generalising Cross Pseudo Supervision to $n$ Networks for Semi-Supervised Semantic Segmentation}

\author{
  Dominik Filipiak$^{1,2} \qquad$
  Piotr Tempczyk$^{1,3} \qquad$
  Marek Cygan$^{3}$
  \\
  $^{1}$AI Clearing, Inc. \\
  $^{2}$Semantic Technology Institute, Department of Computer Science, University of Innsbruck \\
  $^{3}$Institute of Informatics, University of Warsaw
  \\ \tt \{df, pt\}@aiclearing.com, cygan@mimuw.edu.pl
}
\maketitle

\begin{abstract}
  The recent cross pseudo supervision (CPS) approach is a state-of-the-art method for semi-supervised semantic segmentation, which trains two neural networks with a custom cross supervision.
  As we observe that only one of those networks is used in the inference phase, we suggest using both networks using voting and generalising it to more than two networks.
  As a result, we present $n$-CPS, a generalisation of CPS that uses $n$ simultaneously trained subnetworks that learn from each other through one-hot encoding perturbation and consistency regularisation, which together with ensembling of the trained subnetworks significantly improves the performance over the prior method.
  To the best of our knowledge, $n$-CPS paired with CutMix outperforms CPS and sets the new state-of-the-art for Pascal VOC 2012 with (1/16, 1/8, 1/4, and 1/2 supervised regimes) and Cityscapes (1/16 supervised).
  The code is available on GitHub\footnote{The code will be released after the publication. For reviewers, it is available as a supplementary material.}.
\end{abstract}

\input{1_intro.tex}
\input{2_method.tex}
\input{3_evaluation.tex}
\input{4_related_work.tex}

\input{5_summary.tex}
\input{6_acknowledgements.tex}

{\small
\bibliographystyle{ieee_fullname}
\bibliography{egbib}
}

\end{document}

%% file: 1_intro.tex
\section{Introduction}
\label{sec:intro}

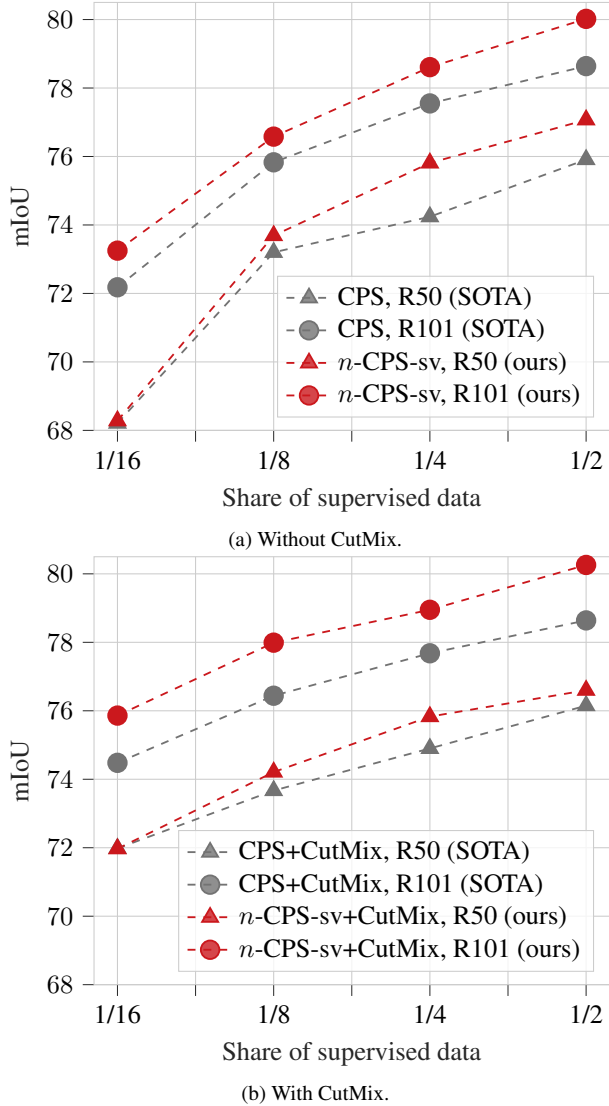
\begin{figure}
  \begin{subfigure}{\linewidth}
    \input{figures/ours_vs_sota.tex}
    \caption{Without CutMix.}
    \label{fig:ours_vs_sota_nocutmix}
  \end{subfigure}
  \begin{subfigure}{\linewidth}
    \input{figures/ours_vs_sota_cutmix.tex}
    \caption{With CutMix.}
    \label{fig:ours_vs_sota_cutmix}
  \end{subfigure}
  \caption{Our method ($n$-CPS-sv, $n=3$) on the Pascal VOC 2012 dataset, compared to the state-of-the-art CPS \protect\cite{chen2021semi}. R50 and R101 denote ResNet-50 and ResNet-101 respectively.}
  \label{fig:ours_vs_sota}
\end{figure}

An intense research effort can be observed in data- and label-efficient machine learning.
The latter can be tackled using semi-supervised learning methods.
Semantic segmentation can significantly benefit from semi-supervised methods due to the relatively high cost of labelling every pixel.
Among numerous techniques, consistency regularisation is proven to boost performance in such settings.
The recent cross pseudo supervision (abbreviated as CPS) approach \cite{chen2021semi} has set the new state-of-the-art in the task of semi-supervised semantic segmentation.
It simultaneously trains two networks, which are penalised for discrepancies between them.
This consistency regularisation mechanism is enriched with a specific one-hot encoding data perturbation.

While in the original CPS approach two networks are used in training, the inference is performed only on the output of the first network.
This observation constituted a question that led us to this study -- why can't we use the outputs of both networks to improve the results? 
This produced another question -- can this approach use more than two networks?
The result of this investigation is $n$-CPS -- a generalisation of the CPS approach.

This paper proposes a generalised cross pseudo supervision approach.
Our contribution is two-fold: 
\begin{itemize}
  \item we generalise the CPS to handle more than two networks in the training process,
  \item we propose an evaluation approach inspired by ensemble learning, which treats these networks similarly to a blend of weak learners to form a better model.
\end{itemize}
To the best of our knowledge, $n$-CPS paired with CutMix \cite{yun2019cutmix} outperforms CPS and sets the new state-of-the-art for Pascal VOC 2012 for 1/16, 1/8, 1/4, and 1/2 supervised regimes and Cityscapes for 1/16 supervised data.
Figure~\ref{fig:ours_vs_sota} presents results for the Pascal VOC 2012 dataset.

The paper is structured as follows. 
In the Section \ref{sec:method}, we present $n$-CPS, the proposed method.
The results of evaluation are presented Section \ref{sec:results} -- the experiment setup in Section \ref{sec:results_setup}, comparison with the state-of-the art in Section \ref{sec:results_results}, and ablation studies in Section \ref{sec:ablations}.
Section \ref{sec:related_work} offers a comprehensive literature review of the literature on the topic.
The paper is concluded with a short summary in Section \ref{sec:summary}.

\begin{figure*}[!h]
  \input{figures/architecture.tikz}
  \caption{The architecture of $n$-CPS during the CPS loss calculation. The slashed line (with //) represents not passing the gradient, whereas the dashed line (- -) denotes cross-pseudo supervision.}
  \label{fig:architecture}
\end{figure*}
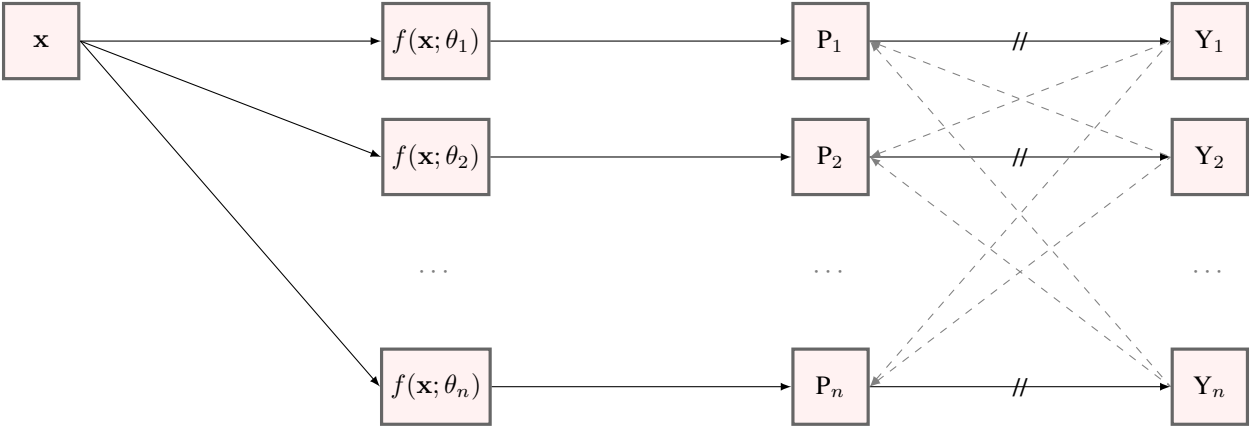

%% file: figures/ours_vs_sota.tex
\begin{tikzpicture}

\definecolor{color0}{rgb}{0.796078431372549,0.0941176470588235,0.113725490196078}

\begin{axis}[
axis line style={white!80!black},
legend cell align={left},
legend style={
  fill opacity=0.8,
  draw opacity=1,
  text opacity=1,
  at={(0.97,0.03)},
  anchor=south east,
  draw=white!80!black
},
tick align=outside,
tick pos=left,
x grid style={white!80!black},
xlabel=\textcolor{white!15!black}{Share of supervised data},
xmajorgrids,
xmin=-0.15, xmax=3.15,
xtick style={color=white!15!black},
xtick={-0.5,0,0.5,1,1.5,2,2.5,3,3.5},
xticklabels={,1/16,,1/8,,1/4,,1/2,},
y grid style={white!80!black},
ylabel=\textcolor{white!15!black}{mIoU},
ymajorgrids,
ymin=68, ymax=80.5,
ytick style={color=white!15!black}
]
\addplot [line width=0.7pt, white!45.0980392156863!black, dashed, mark=triangle*, mark size=3.5, mark options={solid}]
table {%
0 68.21
1 73.2
2 74.24
3 75.91
};
\addlegendentry{CPS, R50 (SOTA)}
\addplot [line width=0.7pt, white!45.0980392156863!black, dashed, mark=*, mark size=3.5, mark options={solid}]
table {%
0 72.18
1 75.83
2 77.55
3 78.64
};
\addlegendentry{CPS, R101 (SOTA)}
\addplot [line width=0.7pt, color0, dashed, mark=triangle*, mark size=3.5, mark options={solid}]
table {%
0 68.28
1 73.69
2 75.81
3 77.07
};
\addlegendentry{$n$-CPS-sv, R50 (ours)}
\addplot [line width=0.7pt, color0, dashed, mark=*, mark size=3.5, mark options={solid}]
table {%
0 73.25
1 76.58
2 78.61
3 80.02
};
\addlegendentry{$n$-CPS-sv, R101 (ours)}
\end{axis}

\end{tikzpicture}

%% file: figures/ours_vs_sota_cutmix.tex
\begin{tikzpicture}

\definecolor{color0}{rgb}{0.796078431372549,0.0941176470588235,0.113725490196078}

\begin{axis}[
axis line style={white!80!black},
legend cell align={left},
legend style={
  fill opacity=0.8,
  draw opacity=1,
  text opacity=1,
  at={(0.97,0.03)},
  anchor=south east,
  draw=white!80!black
},
tick align=outside,
tick pos=left,
x grid style={white!80!black},
xlabel=\textcolor{white!15!black}{Share of supervised data},
xmajorgrids,
xmin=-0.15, xmax=3.15,
xtick style={color=white!15!black},
xtick={-0.5,0,0.5,1,1.5,2,2.5,3,3.5},
xticklabels={,1/16,,1/8,,1/4,,1/2,},
y grid style={white!80!black},
ylabel=\textcolor{white!15!black}{mIoU},
ymajorgrids,
ymin=68, ymax=80.5,
ytick style={color=white!15!black}
]
\addplot [line width=0.7pt, white!45.0980392156863!black, dashed, mark=triangle*, mark size=3.5, mark options={solid}]
table {%
0 71.98
1 73.67
2 74.9
3 76.15
};
\addlegendentry{CPS+CutMix, R50 (SOTA)}
\addplot [line width=0.7pt, white!45.0980392156863!black, dashed, mark=*, mark size=3.5, mark options={solid}]
table {%
0 74.48
1 76.44
2 77.68
3 78.64
};
\addlegendentry{CPS+CutMix, R101 (SOTA)}
\addplot [line width=0.7pt, color0, dashed, mark=triangle*, mark size=3.5, mark options={solid}]
table {%
0 71.97
1 74.21
2 75.83
3 76.6
};
\addlegendentry{$n$-CPS-sv+CutMix, R50 (ours)}
\addplot [line width=0.7pt, color0, dashed, mark=*, mark size=3.5, mark options={solid}]
table {%
0 75.86
1 77.99
2 78.95
3 80.26
};
\addlegendentry{$n$-CPS-sv+CutMix, R101 (ours)}
\end{axis}

\end{tikzpicture}

%% file: figures/architecture.tikz
\tikzset{>=latex}
\begin{tikzpicture}[
  squarednode/.style={rectangle, draw=black!60, fill=red!5, very thick, minimum size=10mm, node distance=0.5cm and 4 cm},
  dotsnode/.style={rectangle, minimum size=10mm, node distance=0.5cm and 4 cm, gray},
  ]
  \node[squarednode] (input) {$\mathbf{x}$};

  \node[squarednode] (n1) [right=of input]             {$f(\mathbf{x}; \theta_1)$};
  \node[squarednode] (n2) [right=of input, below=of n1]{$f(\mathbf{x}; \theta_2)$};
  \node[dotsnode] (n3) [right=of input, below=of n2]{$\cdots$};
  \node[squarednode] (n4) [right=of input, below=of n3]{$f(\mathbf{x}; \theta_n)$};

  \node[squarednode] (p1) [right=of n1]             {$\text{P}_1$};
  \node[squarednode] (p2) [right=of n2, below=of p1]{$\text{P}_2$};
  \node[dotsnode] (p3) [right=of n3, below=of p2]{$\cdots$};
  \node[squarednode] (p4) [right=of n4, below=of p3]{$\text{P}_n$};

  \node[squarednode] (y1) [right=of p1]             {$\text{Y}_1$};
  \node[squarednode] (y2) [right=of p2, below=of y1]{$\text{Y}_2$};
  \node[dotsnode] (y3) [right=of p3, below=of y2]{$\cdots$};
  \node[squarednode] (y4) [right=of p4, below=of y3]{$\text{Y}_n$};

  \draw[->] (input.east) -- (n1.west);
  \draw[->] (input.east) -- (n2.west);
  \draw[->] (input.east) -- (n4.west);

  \draw[->] (n1.east) -- (p1.west);
  \draw[->] (n2.east) -- (p2.west);
  \draw[->] (n4.east) -- (p4.west);

  \draw[->] (p1.east) -- (y1.west) node[pos=0.5,sloped, very thick] {//};
  \draw[->] (p2.east) -- (y2.west) node[pos=0.5,sloped, very thick] {//};
  \draw[->] (p4.east) -- (y4.west) node[pos=0.5,sloped, very thick] {//};

  \draw[<-, dashed, gray] (p1.east) -- (y2.west);
  \draw[<-, dashed, gray] (p1.east) -- (y4.west);
  \draw[<-, dashed, gray] (p2.east) -- (y1.west);
  \draw[<-, dashed, gray] (p2.east) -- (y4.west);
  \draw[<-, dashed, gray] (p4.east) -- (y1.west);
  \draw[<-, dashed, gray] (p4.east) -- (y2.west);

\end{tikzpicture}  

%% file: 2_method.tex
\section{Method}
\label{sec:method}

This section describes the proposed method.
Firstly, we present plain $n$-CPS.
Then, we introduce a variant paired with the CutMix algorithm.
Finally, we present ensemble learning techniques, which increase the evaluation performance.

\paragraph{$\mathbf{n}$-CPS.}
The proposed method generalises the cross pseudo supervision (CPS) approach for semi-supervised semantic segmentation.
Similarly to CPS, a key feature of $n$-CPS is consistency regularisation between the output of one network and the output of another one perturbed with pixel-wise one-hot encoding.
This perturbation (denoted as $\texttt{pmax}$ hereafter) sets 1 for the class with the highest probability and 0 for the others.
However, instead of using two networks (with the same architecture but with differently initialised weights) as in the original CPS algorithm, our method can use $n$ networks (setting $n$ to 2 results in the original CPS approach).
Let $\mathcal{D}_{\text{L}}$ and $\mathcal{D}_{\text{U}}$ denote labelled and unlabelled training data sets respectively, both containing images of size $W \times H$.
The $n$-CPS architecture consists of a set of $n$ networks $f(\mathbf{x}; \theta_j)$ with $j \in \{ 1, \ldots, n \}$, each with the same architecture but initialised with different set of parameters from $\Theta = \{\theta_{1}, \theta_{2}, \ldots, \theta_{n}\}$.
The pixel-wise class probability vector predicted by the $j$-th network $f\left(\mathbf{x}; \theta_j \right)$ on the $i$-th pixel of the image $\mathbf{x}$ is represented by $\mathbf{p}_{ij}$.
The supervised loss is calculated in a standard way:
\begin{equation}
  \mathcal{L}_{\text{L}} = \frac{1}{|\mathcal{D}_{\text{L}}|} \sum_{\mathbf{x} \in \mathcal{D}_{\text{L}}} \frac{1}{W \times H} \sum_{i=1}^{W \times H} \sum_{ j=1}^{n} l(\mathbf{p}_{ij}, \mathbf{y}_{i}^{*}),
\end{equation}
where $l$ is the standard cross-entropy loss function.
The ground truth for the labelled data is marked as $\mathbf{y}_{i}^{*}$.
For the $n$-CPS loss for labelled and unlabelled data, we defined them respectively:
\begin{align}
  \mathcal{L}_{\text{CPS}}^{\text{L}} &= \frac{1}{|\mathcal{D}_{\text{L}}|} \sum_{\mathbf{x} \in \mathcal{D}_{\text{L}}} \frac{1}{W \times H} \sum_{i=1}^{W \times H} \sum_{j=1}^{n} \frac{1}{n-1}\sum_{k\neq j}^{n} \frac{}{} l(\mathbf{p}_{ij}, \mathbf{y}_{ik}), \label{eq:llcps} \\
  \mathcal{L}_{\text{CPS}}^{\text{U}} &= \frac{1}{|\mathcal{D}_{\text{U}}|} \sum_{\mathbf{x} \in \mathcal{D}_{\text{U}}} \frac{1}{W \times H} \sum_{i=1}^{W \times H} \sum_{j=1}^{n} \frac{1}{n-1}\sum_{k\neq j}^{n} \frac{}{} l(\mathbf{p}_{ij}, \mathbf{y}_{ik}). \label{eq:lucps}
\end{align}
This time the loss is calculated with $\mathbf{y}_{ik}$, which is the one-hot encoded output of the $k$-th network on the unlabelled data.
Finally, the overall loss $\mathcal{L}$ is defined as:
\begin{equation}
  \mathcal{L} = \mathcal{L}_{\text{L}} + \lambda \left( \mathcal{L}_{\text{CPS}}^{\text{L}} + \mathcal{L}_{\text{CPS}}^{\text{U}} \right),
\end{equation}
where $\lambda$ is the weight of CPS loss.
In equations \eqref{eq:llcps} and \eqref{eq:lucps} we scale the consistency loss for each pair by $1/(n-1) $ to maintain a balance between supervised and CPS part of the loss regardless of the value of $n$.
This scaling factor has this form because each of the $n$ networks in our algorithm can be paired with $n-1$ other networks to calculate CPS loss.
Figure \ref{fig:architecture} depicts calculating the CPS loss.

\begin{algorithm}
  \caption{$n$-CPS}
  \label{alg:ncps}
  \begin{algorithmic}
  \FOR {$\textbf{each} \text{ batch}$}
  \STATE $\mathcal{L} \leftarrow 0, \quad \mathcal{L}_{\text{L}} \leftarrow 0, \quad \mathcal{L}_{\text{CPS}}^{\text{L}} \leftarrow 0, \quad \mathcal{L}_{\text{CPS}}^{\text{U}} \leftarrow 0 $
  \STATE $ \mathbf{x}^{L}, \mathbf{x}^{U}, \mathbf{Y}^{*} \leftarrow \texttt{dataloader.iter()}$
  \FOR {$j=1,\ldots,n$}
    \STATE $\mathbf{P}^{L}_{j} \leftarrow f(\mathbf{x}^{L}; \theta_{j}) $
    \STATE $\mathbf{P}^{U}_{j} \leftarrow f(\mathbf{x}^{U}; \theta_{j})$ 
  \ENDFOR
  \FOR {$(l,r) \in \left\{(l,r) \in \{1,\ldots,n\}^2 : l < r \right\}$}
    \STATE $\mathbf{Y}_{l}^{U} \leftarrow \texttt{pmax}(\mathbf{P}^{U}_{l})$ \COMMENT{\cancel{$\nabla$}}
    \STATE $\mathbf{Y}_{r}^{U} \leftarrow \texttt{pmax}(\mathbf{P}^{U}_{r})$ \COMMENT{\cancel{$\nabla$}}
    \STATE $\mathbf{Y}_{l}^{L} \leftarrow \texttt{pmax}(\mathbf{P}^{L}_{l})$ \COMMENT{\cancel{$\nabla$}}
    \STATE $\mathbf{Y}_{r}^{L} \leftarrow \texttt{pmax}(\mathbf{P}^{L}_{r})$ \COMMENT{\cancel{$\nabla$}}
    \STATE $\mathcal{L}_{\text{CPS}}^{\text{U}} \leftarrow \mathcal{L}_{\text{CPS}}^{\text{U}} + \mathcal{L}_{\text{CPS}} \left( \mathbf{P}^{U}_{l}, \mathbf{Y}^{U}_{r} \right) + \mathcal{L}_{\text{CPS}} \left( \mathbf{P}^{U}_{r}, \mathbf{Y}^{U}_{l} \right)$
    \STATE $\mathcal{L}_{\text{CPS}}^{\text{L}} \leftarrow \mathcal{L}_{\text{CPS}}^{\text{L}} + \mathcal{L}_{\text{CPS}} \left( \mathbf{P}^{L}_{l}, \mathbf{Y}^{L}_{r} \right) + \mathcal{L}_{\text{CPS}} \left( \mathbf{P}^{L}_{r}, \mathbf{Y}^{L}_{l} \right)$
  \ENDFOR
  \FOR {$j=1,\ldots,n$}
    \STATE $\mathcal{L}_{\text{L}} \leftarrow \mathcal{L}_{\text{L}} + \mathcal{L}_{\text{L}}\left(\mathbf{P}^{L}_{j}, \mathbf{Y}^{*} \right)$
  \ENDFOR
  \STATE $\mathcal{L} \leftarrow \mathcal{L}_{\text{L}} + \frac{\lambda}{(n-1)} \left( \mathcal{L}_{\text{CPS}}^{\text{U}} + \mathcal{L}_{\text{CPS}}^{\text{L}} \right)$
  \STATE $\mathcal{L}.\texttt{backward()}$
  \ENDFOR
  \end{algorithmic}
  \end{algorithm}

Details are presented in Algorithm \ref{alg:ncps}.
In each batch, we select images from the labelled and unlabelled data set, which are denoted as $\mathbf{x}^{L}, \mathbf{x}^{U}$ respectively.
Then, we perform the forward passes on all the data using all the networks separately.
The $n$-CPS model consists of $n$ networks, and the output of each network is the pixel-wise probability of each class.
The $i$-th one is represented by $f(\ \cdot\ ; \theta_{i})$.
Then, we perform the cross pseudo supervision for each unique pair of networks $\binom{n}{2}$ times.
The one-hot encoded outputs of the $l$-th and $r$-th networks are denoted as $\mathbf{Y}_{l}$ and $\mathbf{Y}_{r}$.
In this case, $\texttt{pmax}$ is a pixel-wise maximum function that selects each pixel's maximum value from the two masked images.
It is calculated without passing the gradient (denoted as \cancel{$\nabla$} in the algorithm).
The CPS loss is performed both on the labelled and unlabelled data.
Then, we calculate the standard supervised loss for each network.
The overall loss consists of the standard labelled loss $\mathcal{L}_{\text{L}}$ and the CPS losses $\mathcal{L}_{\text{CPS}}^{\text{L}}$ and $\mathcal{L}_{\text{CPS}}^{\text{U}}$ (for labelled and unlabelled data weighted by $\lambda$ and normalised by the factor of $\frac{1}{n-1}$).
While $n$-CPS needs $\binom{n}{2}$ CPS loss calculations per batch, the most computationally expensive forward calls are hit only $2n$ times per batch, which \emph{effectively} makes the time complexity of the algorithm linear in $n$.

\paragraph{$\mathbf{n}$-CPS with CutMix.}
We also used the CutMix augmentation algorithm \cite{yun2019cutmix} adapted to semantic segmentation \cite{french2019semi}, as it was proven to be very effective in the original CPS approach \cite{chen2021semi}.
Algorithm \ref{alg:ncps-cutmix} presents our method.
In each batch, we select images from the labelled and unlabelled data set, and we randomly select a binary mask $M$.
Regarding the data, $\mathbf{x}^{L}, \mathbf{x}^{U}, \mathbf{x}^{m}$ represent the labelled, unlabelled and mixed data respectively.
Notice that there are two different subsets of unlabelled data per batch -- $\mathbf{x}^{U}_{1}$ and $\mathbf{x}^{U}_{2}$.
They are sampled differently from the labelled data and both are used to generate the mixed data $\mathbf{x}^{m}$, which is the result of the CutMix applied with batch-wise mask $\mathbf{M}$.
The CutMix algorithm is defined as follows:
\begin{equation}
  \texttt{CutMix}(\mathbf{x}^{U}_1, \mathbf{x}^{U}_2, \textbf{M}) = \left( 1 - \textbf{M} \right) \odot \mathbf{x}^{U}_1 + \textbf{M} \odot \mathbf{x}^{U}_2.
\end{equation}
The mask is the same size as the image and is randomly generated to satisfy CutMix constraints (such as leaving out the rectangular area).

Then, we perform the forward passes on all the data using all the networks separately, similarly to $n$-CPS.
Cross pseudo supervision for each unique pair of networks is performed slightly differently regarding one-hot encoding vectors.
They are obtained by combining images in a CutMix-like approach and applying the maximum function.
The overall loss consists of the standard labelled loss $\mathcal{L}_{\text{L}}$ and the CPS loss $\mathcal{L}_{\text{CPS}}^{\text{U}}$ weighted by $\lambda$ and normalised by the factor of $\frac{1}{n-1}$.
Similarly to the original CPS with CutMix augmentation, the CPS loss is not calculated on supervised data and therefore $\mathcal{L}_{\text{CPS}}^{\text{L}}$ is not calculated.
Once again, forward calls are linear in $n$ and setting $n=2$ results in the original CPS+CutMix approach.

\begin{algorithm}
  \caption{$n$-CPS+CutMix}
  \label{alg:ncps-cutmix}
  \begin{algorithmic}
  \FOR {$\textbf{each} \text{ batch}$}
  \STATE $\mathcal{L} \leftarrow 0, \quad \mathcal{L}_{\text{L}} \leftarrow 0, \quad \mathcal{L}_{\text{CPS}}^{\text{U}} \leftarrow 0 $
  \STATE $ \mathbf{x}^{L}, \mathbf{x}^{U}_{1}, \mathbf{x}^{U}_{2}, \textbf{M}, \mathbf{Y}^{*} \leftarrow \texttt{dataloader.iter()}$
  \STATE $\mathbf{x}^{m} \leftarrow \texttt{CutMix}(\mathbf{x}^{U}_1, \mathbf{x}^{U}_2, \textbf{M})$
  \FOR {$j=1,\ldots,n$}
    \STATE $\mathbf{P}^{L}_{j} \leftarrow   f(\mathbf{x}^{L}; \theta_{j}) $
    \STATE $\mathbf{P}^{m}_{j} \leftarrow   f(\mathbf{x}^{m}; \theta_{j}) $
    \STATE $\mathbf{P}^{U}_{j,1} \leftarrow f(\mathbf{x}^{U}_1; \theta_{j})$
    \STATE $\mathbf{P}^{U}_{j,2} \leftarrow f(\mathbf{x}^{U}_2; \theta_{j}) $
  \ENDFOR
  \FOR {$(l,r) \in \left\{(l,r) \in \{1,\ldots,n\}^2 : l < r \right\}$}
    \STATE $\mathbf{Y}_{l} \leftarrow \texttt{pmax}(\mathbf{P}^{U}_{l,1} \odot (1 - \mathbf{M}) + \mathbf{P}^{U}_{l,2} \odot \mathbf{M} ) $ \COMMENT{\cancel{$\nabla$}}
    \STATE $\mathbf{Y}_{r} \leftarrow \texttt{pmax}(\mathbf{P}^{U}_{r,1} \odot (1 - \mathbf{M}) + \mathbf{P}^{U}_{r,2} \odot \mathbf{M} ) $ \COMMENT{\cancel{$\nabla$}} 
    \STATE $\mathcal{L}_{\text{CPS}}^{\text{U}} \leftarrow \mathcal{L}_{\text{CPS}}^{\text{U}} + \mathcal{L}_{\text{CPS}} \left( \mathbf{P}^{m}_{l}, \mathbf{Y}_{r} \right) + \mathcal{L}_{\text{CPS}} \left( \mathbf{P}^{m}_{r}, \mathbf{Y}_{l} \right)$
  \ENDFOR
  \FOR {$j=1,\ldots,n$}
    \STATE $\mathcal{L}_{\text{L}} \leftarrow \mathcal{L}_{\text{L}} + \mathcal{L}_{\text{L}}\left(\mathbf{P}^{L}_{j}, \mathbf{Y}^{*} \right)$
  \ENDFOR
  \STATE $\mathcal{L} \leftarrow \mathcal{L}_{\text{L}} + \frac{\lambda}{(n-1)} \mathcal{L}_{\text{CPS}}^{\text{U}}$
  \STATE $\mathcal{L}.\texttt{backward()}$
  \ENDFOR
  \end{algorithmic}
  \end{algorithm}

\paragraph{Ensemble learning techniques.}
In the language of ensemble learning, an \emph{ensemble} is a set of weak learners (models trained independently, each of them with relatively low performance), which together forms a strong model (which should display better performance) \cite{rokach2010ensemble}.
During the $n$-CPS training process, there are $n$ networks trained separately.
In the original approach, the evaluation used only the results of the first network and discarded the other ones.
While this approach alone was proven to generate state-of-the-art results, our empirical results show that including all the information from the trained networks is beneficial for performance.
In the original CPS paper, the experiment showed that even in the last steps of learning, approximately 5\% of pixels are labelled differently by the trained networks.

Therefore, we use the results of all the networks.
It can be realised by taking the pixel-wise softmax of the output of each network and then combining the results in several ways.
In this paper, we test two of them: max confidence (denoted as \texttt{mc}) and soft voting (\texttt{sv}).
In the max confidence approach, we choose the result with the highest score.
In other words, for each pixel, we choose the class from the most \emph{confident} network.
In the PyTorch-like notation, that max confidence voting is defined as:
\texttt{softmax(y, dim=2).max(dim=0)}, where the concatenated output of all networks \texttt{y} is of shape \texttt{(n, b, c, w, h)} -- representing consecutively number of the networks, batch size, number of classes, width and height.
The soft voting approach is similar, but instead of choosing the class with the highest score, we sum the probabilities of all the networks.
This is proportional to the weighted mean of the output of all networks.
In the PyTorch-like notation that would translate to \texttt{softmax(y, dim=2).sum(dim=0)}.
While the networks are not necessarily weak learners in the ensemble learning theory sense, both of the approaches are proven to be effective and improve the performance of $n$-CPS (see sections \ref{sec:results_results} and \ref{sec:ablations}).

%% file: 3_evaluation.tex
\section{Evaluation}
\label{sec:results}

This section describes the evaluation of the proposed approach.
First, we present the setup of our experiments and the training details.
We evaluate $n$-CPS on PASCAL VOC 2012 and Cityscapes data sets.
Ablation studies are also performed.

\subsection{Experiment setup}
\label{sec:results_setup}

\paragraph{Datasets.}
The datasets used in this experiment are the same as in the original CPS paper \cite{chen2021semi} -- PASCAL VOC 2012 \cite{everingham2010pascal} and Cityscapes \cite{cordts2016cityscapes}.
The PASCAL VOC 2012 contains 21 classes (including the background class). 
Regarding Cityscapes, it comprises 30 classes.
We also follow GCT \cite{ke2020guided} protocols regarding the ratio of supervised-to-unsupervised images in the dataset (1/16, 1/8, 1/4, 1/2).
The sampling scheme is taken from the CPS paper \cite{chen2021semi} to provide a fair comparison with other methods.
Both data sets are evaluated using the standard mean intersection over union (mIoU) metric in \% over \texttt{val} sets (1,456 images in PASCAL VOC 2012 and 500 in Cityscapes).

\paragraph{Training details.}
We use the same training setup as in the original CPS paper \cite{chen2021semi}.
We extend the original PyTorch codebase with our $n$-CPS approach, although training details (such as augmentations or hyperparameters) stays the same. 
This means that we use DeepLabv3+ \cite{deeplabv3plus2018} as our network with ResNet-50/101 backbones \cite{he2016deep} paired with mini-batch SGD with momentum ($0.9$) and weight decay ($0.0005$).
Similarly to the authors of CPS, we also use the poly learning rate policy, in which the initial learning rate is multiplied by $\left( 1 - \frac{\texttt{iter}}{\texttt{max\_iter}} \right)^{0.9}$.
For Pascal VOC, both supervised and CPS loss is computed using standard cross-entropy loss.
Regarding Cityscapes, OHEM loss \cite{shrivastava2016training} is used for supervised loss and cross-entropy loss for CPS loss.
VOC models were trained on 4$\times$V100 GPUs, whereas Cityscapes models on 8$\times$V100 GPUs.

\subsection{Results}
\label{sec:results_results}

We report mIoU results from the network with the highest-scoring step (not necessarily the last one).
These results use $n=3$ as the number of networks and \texttt{mc} and \texttt{sv} ensembling techniques\footnote{Our evaluation was primarily meant to be done on max confidence voting. 
Soft voting was added later and was tested only on the models that were best performing with the max confidence voting.
This means that there is a chance that reported soft voting results might be slightly improved if tested on all steps.}.
The training regime is taken from the CPS paper, and it consists of different supervision ratios (1/16, 1/8, 1/4, 1/2) trained for 32/34/40/60 and 128/137/160/240 steps for Pascal VOC and Cityscapes, respectively.
We report the best test results during the evaluation (not necessarily the result from the last step).

\begin{figure}[!htpb]
  \includegraphics[width=\linewidth]{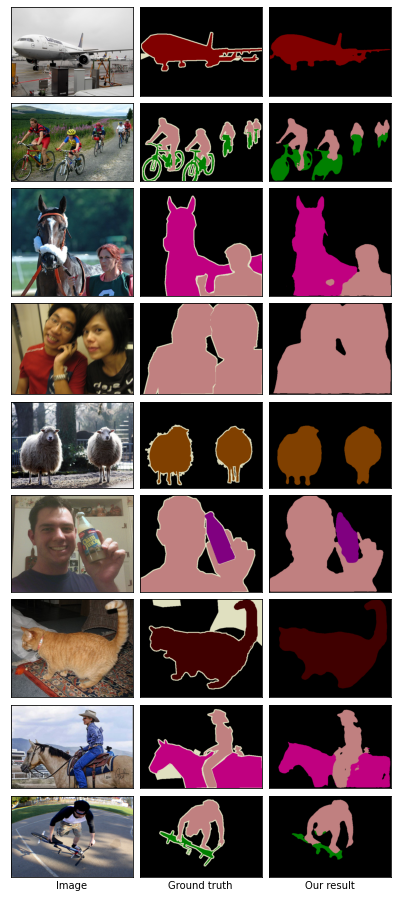}
  \caption{Example results for $3$-CPS-sv (DeepLabv3+, ResNet50, 1/8 supervision, no CutMix) on Pascal VOC 2012 \texttt{val} dataset.}
  \label{fig:examples}
\end{figure}

\paragraph{Pascal VOC 2012.}
The first part of Table \ref{tab:results_voc} presents the results on the Pascal VOC dataset compared to other recent methods (the non-our results are taken from the CPS paper \cite{chen2021semi}).
For different supervision regimes (1/16, 1/8, 1/4, 1/2), the version without CutMix outperforms the mIoU results reported in the original CPS paper (previous state-of-the-art) by $+0.15$/$+0.49$/$+1.57$/$+1.16$ percentage points for ResNet-50 and $+1.33$/$+0.75$/$+1.06$/$+1.38$ pp for ResNet-101.
Regarding the version with CutMix algorithm, the $n$-CPS approach is better than CPS by $+0.05$/$+0.54$/$+0.95$/$+0.50$ pp for ResNet-50 and $+1.38$/$+1.55$/$+1.29$/$+1.62$ pp for ResNet-101.
Mean confidence and soft voting ensembles behave similarly, and usually, the difference is slight (no more than $\pm 0.1$ mIoU). 
Figure \ref{fig:examples} shows masks from the 3-CPS model on this dataset.

Interestingly, for ResNet-50, the version without CutMix outperforms the version with CutMix in the 1/2 supervised scenario.
The observed trend suggests that CutMix is vital for small supervision ratios, but there is a visible decrease in its effectiveness with the increasing share of supervised data.
Since CutMix can be perceived as a heavy augmentation technique, we hypothesise that it can even deteriorate the training for higher shares of supervised data.
However, this effect is not observed with ResNet-101 -- at least up to 1/2 supervision scheme.

\begin{table*}[htpb]
  \caption{Comparison of the semi-supervised segmentation methods (mIoU under different supervision regimes, DeepLabv3+). 
  }
  \centering
  \begin{tabular}{@{}lrrrrrrrr@{}}
  \toprule
  \multicolumn{1}{c}{\multirow{2}{*}{}} & \multicolumn{4}{c}{ResNet-50} & \multicolumn{4}{c}{ResNet-101} \\ \cmidrule(lr){2-5} \cmidrule(lr){6-9}
  \multicolumn{1}{c}{} &
  \multicolumn{1}{c}{1/16} &
  \multicolumn{1}{c}{1/8} &
  \multicolumn{1}{c}{1/4} &
  \multicolumn{1}{c}{1/2} &
  \multicolumn{1}{c}{1/16} &
  \multicolumn{1}{c}{1/8} &
  \multicolumn{1}{c}{1/4} &
  \multicolumn{1}{c}{1/2}  \\ \midrule
  \multicolumn{9}{c}{\emph{Pascal VOC 2012}} \\ \midrule
  MT \cite{tarvainen2017mean} & 66.77 & 70.78 & 73.22 & 75.41 & 70.59 & 73.20 & 76.62 & 77.61 \\
  CCT \cite{ouali2020semi} & 65.22 & 70.87 & 73.43 & 74.75 & 67.94 & 73.00 & 76.17 & 77.56 \\
  CutMix-Seg \cite{french2019semi} & 68.90 & 70.70 & 72.46 & 74.49 & 72.56 & 72.69 & 74.25 & 75.89 \\
  GCT \cite{ke2020guided} & 64.05 & 70.47 & 73.45 & 75.20 & 69.77 & 73.30 & 75.25 & 77.14 \\
  CPS \cite{chen2021semi} & 68.21 & 73.20  & 74.24 & 75.91 & 72.18  & 75.83 & 77.55 & 78.64 \\
  CPS+CutMix \cite{chen2021semi} & 71.98 & 73.67 & 74.90  & 76.15 & 74.48  & 76.44 & 77.68 & 78.64 \\ \midrule
  3-CPS-mc (ours) & 68.36 & 73.45 & 75.75 & 77.00 & 73.51 & 76.46 & 78.59 & 79.90 \\
  3-CPS-sv (ours) & 68.28 & 73.69 & 75.81 & \textbf{77.07} & 73.25 & 76.58 & 78.61 & 80.02\\ \cmidrule(l){2-9}
  3-CPS-mc+CutMix (ours) & \textbf{72.03} & 74.18 & \textbf{75.85} & 76.65 & 75.80 & 77.96 & \textbf{78.97} & 80.06 \\
  3-CPS-sv+CutMix (ours) & 71.97 & \textbf{74.21} & 75.83 & 76.6 & \textbf{75.86} & \textbf{77.99} & 78.95 & \textbf{80.26} \\
  \midrule
  \multicolumn{9}{c}{\emph{Cityscapes}} \\ \midrule
  MT \cite{tarvainen2017mean} & 66.14 & 72.03 & 74.47 & 77.43 & 68.08 & 73.71 & 76.53 & 78.59 \\
  CCT \cite{ouali2020semi} & 66.35 & 72.46 & 75.68 & 76.78 & 69.64 & 74.48 & 76.35 & 78.29 \\
  GCT \cite{ke2020guided} & 65.81 & 71.33 & 75.30 & 77.09 & 66.90 & 72.96 & 76.45 & 78.58 \\
  CPS \cite{chen2021semi} & 69.79 & 74.39 & 76.85 & 78.64 & 70.50 & 75.71 & 77.41 & 80.08 \\
  CPS+CutMix \cite{chen2021semi} & 74.47 & 76.61 & 77.83 & 78.77 & 74.72 & 77.62 & 79.21 & 80.21 \\ \midrule
  3-CPS-mc (ours) & 69.78 & 74.80 & 76.74	& \textbf{79.29} & -- & -- & -- & -- \\ 
  3-CPS-sv (ours) & 69.76 & 74.88 & 76.74 & 79.27 & -- & -- & -- & -- \\ \cmidrule(l){2-9}
  3-CPS-mc+CutMix (ours) & 76.06 & 77.58 & 78.36 & 79.15 & -- & -- & -- & --\\
  3-CPS-sv+CutMix (ours) & \textbf{76.08} & \textbf{77.61} & \textbf{78.41} & 79.2 & -- & -- & -- & -- \\
  \bottomrule
  \end{tabular}
  \label{tab:results_voc}
\end{table*}

\paragraph{Cityscapes.}

The second part of Table \ref{tab:results_voc} presents the results on the Cityscapes dataset compared with other recent methods.
Similarly, the non-our results are taken from the CPS paper \protect\cite{chen2021semi}.
We do not report results for the ResNet-101 backbone network due to the unavailability of the appropriate hardware (8$\times$V100 GPUs were not enough in terms of RAM).
On ResNet-50 and without CutMix, $n$-CPS achieved $-0.01$/$+0.49$/$-0.11$/$+0.65$ pp change of mIoU compared to CPS.
With CutMix on, our approach outperforms the current state of the art by $+1.61$/$+1.00$/$+0.58$/$+0.43$ pp.
Even without testing our approach on ResNet-101, the $n$-CPS on ResNet-50 outperforms CPS on ResNet-101 on 1/16 supervision ($+1.36$ mIoU).
Interestingly, the model based on ResNet-50 with the CutMix algorithm showed a slightly worse performance 1/2 supervision than the model without it.
This behaviour is consistent with the PASCAL VOC 2012 results and possibly it can share a similar explanation.

\paragraph{Variance of the results.}

Results reported in Table \ref{tab:results_voc} are collected from single runs.
We decided to run the experiment several times to assess the method's variance and collect the results.
Table \ref{table:n3-voc8-resnet50_multiple_runs} shows the results of such experiment for $n$-CPS ($n=3, \lambda=1.5$) on Pascal VOC 1/8 supervised (ResNet-50, without CutMix) with five runs.
We report the best models for single network inference, max confidence, and soft voting.
The results are slightly different from the results reported in Table \ref{tab:results_voc} and Table \ref{tab:results_ablation_ncps}, as they were collected in a separate run.
All the methods have their standard deviation less than $0.25$ pp.
The means for ensembling-based evaluation are almost identical, although the standard deviations are slightly smaller for 3-CPS-sv.

\input{tables/n3-voc8-resnet50_multiple_runs.tex}

\subsection{Ablations}
\label{sec:ablations}

This section presents the ablation experiments on the Pascal VOC 2012 dataset.
We try to assess the importance of generalising CPS by controlling the number of networks $n$.
Then, we address the influence of the max confidence ensembling technique without cross-pseudo supervision.
Finally, we study how the choice of ensembling technique influences the performance.

\paragraph{Importance of $n$-CPS.}

To assess the importance of $n$-CPS, we run a series of experiments on the PASCAL VOC 2012 dataset with different values.
The results are reported in Table \ref{tab:results_ablation_ncps} in terms of the mIoU under different supervision regimes.
We report the results of the original CPS \cite{chen2021semi}, as well as the results of our implementation ($n$-CPS, where $n \in \{2, 3\}$).
All networks were trained on DeepLabv3+ with ResNet-50/101 as a backbone and using $\lambda=1.5$. 
Most importantly, we did not use any ensembling techniques here.
Note that while the original CPS is equivalent to ours reproduced with $2$-CPS, these results are slightly different due to randomness.
Nevertheless, one can observe that evaluating only the first network $f_1$ (as in the original CPS paper), which is not necessarily the best performing one, does not show the immediate advantage of using $n$-CPS (except ResNet-101 without CutMix).

\paragraph{Importance of ensemble networks.}

The previous paragraph showed that the choice of increased $n$ does not improve the performance alone.
However, in Section \ref{sec:results_results} we have already shown that the method combined with proper ensembling technique yields state-of-the-art results.
To isolate the effect of ensembling, we performed a series of experiments with $\lambda$ set to zero.
This disabled cross-pseudo supervision loss from the learning process and effectively turned the task to standard supervised ensemble learning.
Table \ref{tab:results_ablation_ensemble} presents the results of this ablation study.
As it turns out, for 2-CPS, ensembling resulted in $+1.57$ pp to mIoU, while ensembling and CPS ($\lambda=1.5$) provided another $+2.78$ pp for 1/16 supervised data.
The trend is consistent in different supervision regimes, though it diminishes with larger shares of supervised data ($+1.47$/$+2.04$ pp for 1/8 supervised, $+1.13$/$+1.44$ pp for 1/4 supervised, $+0.59$/$+0.42$ pp for 1/2 supervised).
Regarding 3-CPS, one can observe similar behaviour ($+2.09$ pp with ensembling and $+2.08$ pp  with $\lambda=1.5$ for 1/16 supervised, $+1.79$/$+1.88$ pp for 1/8 supervised, $+2.04$/$+0.72$ pp for 1/4 supervised, $+1.53$/$+0.02$ pp for 1/2 supervised).
This shows that the choice of ensembling and CPS positively affects the performance, especially compared to the ensembling alone.
Regarding the $n$ with ensembling, we observe that the performance of the 3-CPS-ens is better than 2-CPS ($+0.11$ pp for 1/8 supervised, $+0.10$ pp for 1/4 supervised, $+0.27$ pp for 1/2 supervised) except 1/16 supervised ($-0.60$ pp).
This justifies the generalisation of the original CPS combined with ensembling for larger shares of supervised data.

\begin{table*}[]
  \caption{Ablation study on the number of networks (no ensemble techniques, DeepLabv3+ with ResNet-50/101 backbone, $\lambda = 1.5$).}
  \centering
  \begin{tabular}{@{}lrrrrrrrr@{}}
  \toprule
  \multicolumn{1}{c}{\multirow{2}{*}{}} & \multicolumn{4}{c}{ResNet-50} & \multicolumn{4}{c}{ResNet-101} \\ \cmidrule(lr){2-5} \cmidrule(lr){6-9}
  \multicolumn{1}{c}{} &
  \multicolumn{1}{c}{1/16} &
  \multicolumn{1}{c}{1/8} &
  \multicolumn{1}{c}{1/4} &
  \multicolumn{1}{c}{1/2} &
  \multicolumn{1}{c}{1/16} &
  \multicolumn{1}{c}{1/8} &
  \multicolumn{1}{c}{1/4} &
  \multicolumn{1}{c}{1/2}  \\ \midrule
  CPS (Chen \emph{et al.}) & 68.21 & 73.20 & 74.24 & 75.91 &72.18 & 75.83 & 77.55 & 78.64\\
  2-CPS (ours) & 68.64 & 73.30 & 75.34 & 76.33 &73.15 & 76.07 & 77.32 & 78.64\\
  3-CPS (ours) & 67.5 & 72.67 & 75.12 & 76.36 & 73.32 & 76.03 & 78.25 & 78.87\\ \cmidrule{2-9}
  CPS+CutMix (Chen \emph{et al.}) &  71.98 & 73.67 & 74.90  & 76.15 & 74.48 & 76.44 & 77.68 & 78.64\\
  2-CPS+CutMix (ours) & 71.43 & 73.99 & 75.37 & 75.6 & 74.59 & 77.11 & 77.64 & 78.65\\
  3-CPS+CutMix (ours) & 71.11 & 73.56 & 74.68 & 75.86 & 74.98 & 76.98 & 77.95 & 79.67 \\
\bottomrule
\end{tabular}
\label{tab:results_ablation_ncps}
\end{table*}

\begin{table}[]
  \caption{Ablation study on importance of CPS ($\lambda \in \{ 0, 1.5\}$), $n$-CPS ($n \in \{2, 3\}$) and ensembling (none, max confidence) on Pascal VOC (ResNet-50, without CutMix).}
  \centering
  \begin{tabular}{@{}lrrrrr@{}}
  \toprule
  \multicolumn{1}{c}{} &
  \multicolumn{1}{c}{$\lambda$} &
  \multicolumn{1}{c}{1/16} &
  \multicolumn{1}{c}{1/8} &
  \multicolumn{1}{c}{1/4} &
  \multicolumn{1}{c}{1/2} \\ \midrule
2-CPS & 0 & 64.61 & 69.83 & 73.08 & 75.72 \\
2-CPS-mc & 0 & 66.18 & 71.3 & 74.21 & 76.13 \\ \cmidrule{3-6} 
2-CPS-mc & 1.5 & 68.96	& 73.34	& 75.65	& 76.73 \\ \cmidrule{2-6}
3-CPS & 0 & 64.19 & 69.78 & 72.63 & 75.45 \\
3-CPS-mc & 0 & 66.28 & 71.57 & 75.03 & 76.98 \\ \cmidrule{3-6}
3-CPS-mc & 1.5 & 68.36 & 73.45 & 75.75 & 77.00 \\
\bottomrule
\end{tabular}
\label{tab:results_ablation_ensemble}
\end{table}

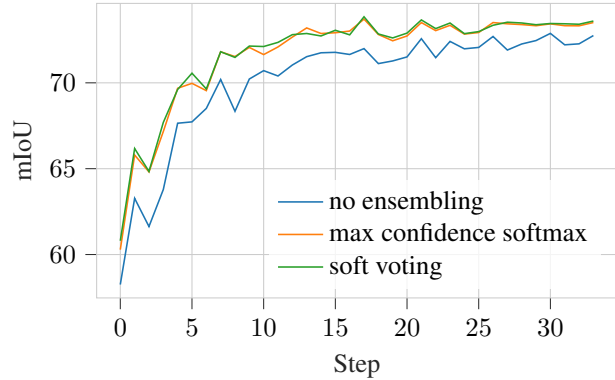
\begin{figure}
    \input{figures/ablation-3cps-voc8-ensembling}
    \caption{Ablation study of different ensembling methods for 3-CPS on Pascal VOC 1/8 supervised (ResNet-50, without CutMix).}
    \label{fig:ablation-3cps-voc8-ensembling}
\end{figure}

\paragraph{Importance of the ensemble learning method.}
We also investigate how the choice of ensembling technique influences the performance during the whole training process.
Apart from the best results reported in Table \ref{tab:results_voc}, we also control it in a whole training procedure.
To do so, we trained $n$-CPS ($n=3, \lambda=1.5$) with the 1/8 supervised regime with ResNet-50 on the Pascal VOC dataset and control evaluation results for different types of ensembling techniques presented in Section \ref{sec:method}: no ensembling (i.e. only the first network is evaluated), max confidence and soft voting.
Figure \ref{fig:ablation-3cps-voc8-ensembling} shows the results of the ablation study.
The highest mIoU scores were $73.85$ for soft voting and $73.71$ for max confidence.
In this case, soft voting was better than max confidence by $0.11$ mIoU points on average.
Soft voting was also better on the majority of learning steps.
Notice that these results are slightly different from the results reported in Table \ref{tab:results_voc}, as they were collected in a separate run.

%% file: tables/n3-voc8-resnet50_multiple_runs.tex
\begin{table*}
\centering
\caption{Statistics of the multiple runs of the same training (3-CPS, 1/8 supervised, ResNet50 on Pascal VOC, $\lambda=1.5$).}
\label{table:n3-voc8-resnet50_multiple_runs}
\begin{tabular}{lrrrrrrrrr}
\toprule
{} & \multicolumn{5}{c}{run} & \multirow{2}{*}{mean} & \multirow{2}{*}{std}& \multirow{2}{*}{min}& \multirow{2}{*}{max} \\ \cmidrule(lr){2-6}
{} & \#1 & \#2 & \#3 & \#4 & \#5 & & &  &  \\
\midrule
3-CPS    & 72.91 & 72.57 & 72.94 & 72.75 & 73.17 & 72.87 &  0.23 &  72.57 &  73.17 \\
3-CPS-mc & 73.54 & 73.42 & 73.60 & 73.16 & 73.55 & 73.45 &  0.18 &  73.16 &  73.60 \\
3-CPS-sv & 73.50 & 73.40 & 73.62 & 73.22 & 73.50 & 73.45 &  0.15 &  73.22 &  73.62 \\
\bottomrule
\end{tabular}
\end{table*}

%% file: figures/ablation-3cps-voc8-ensembling.tex
\begin{tikzpicture}

\definecolor{color0}{rgb}{0.12156862745098,0.466666666666667,0.705882352941177}
\definecolor{color1}{rgb}{1,0.498039215686275,0.0549019607843137}
\definecolor{color2}{rgb}{0.172549019607843,0.627450980392157,0.172549019607843}

\begin{axis}[
axis line style={white!80!black},
height=5.5cm,
legend cell align={left},
legend style={
  fill opacity=0.8,
  draw opacity=1,
  text opacity=1,
  at={(0.97,0.03)},
  anchor=south east,
  draw=none
},
tick align=outside,
tick pos=left,
width=8.5cm,
x grid style={white!80!black},
xlabel=\textcolor{white!15!black}{Step},
xmajorgrids,
xmin=-1.65, xmax=34.65,
xtick style={color=white!15!black},
y grid style={white!80!black},
ylabel=\textcolor{white!15!black}{mIoU},
ymajorgrids,
ymin=57.473895072937, ymax=74.6263360977173,
ytick style={color=white!15!black}
]
\addplot [semithick, color0]
table {%
0 58.2535514831543
1 63.2734794616699
2 61.6302833557129
3 63.7861785888672
4 67.6482849121094
5 67.7262268066406
6 68.5099105834961
7 70.1954193115234
8 68.3437805175781
9 70.2226181030273
10 70.7097549438477
11 70.3991317749023
12 71.0368728637695
13 71.5234680175781
14 71.750114440918
15 71.7761764526367
16 71.6518096923828
17 71.9936599731445
18 71.1208724975586
19 71.2751312255859
20 71.508903503418
21 72.5703048706055
22 71.4660720825195
23 72.4124450683594
24 71.9816818237305
25 72.0623931884766
26 72.7006072998047
27 71.9096908569336
28 72.2627487182617
29 72.4620819091797
30 72.8799514770508
31 72.2140502929688
32 72.2711410522461
33 72.7528610229492
};
\addlegendentry{no ensembling}
\addplot [semithick, color1]
table {%
0 60.2847785949707
1 65.7940444946289
2 64.8131408691406
3 67.1637954711914
4 69.6831283569336
5 69.9705047607422
6 69.5455703735352
7 71.8090744018555
8 71.5299377441406
9 72.0623779296875
10 71.6453552246094
11 72.0901794433594
12 72.6637191772461
13 73.1973724365234
14 72.8734130859375
15 72.9169616699219
16 73.0159072875977
17 73.7104110717773
18 72.8152770996094
19 72.4487533569336
20 72.7292175292969
21 73.5112075805664
22 73.0403442382812
23 73.3496398925781
24 72.8379440307617
25 72.9300689697266
26 73.5109329223633
27 73.4370269775391
28 73.391975402832
29 73.3255844116211
30 73.4334716796875
31 73.3277206420898
32 73.3207778930664
33 73.5121765136719
};
\addlegendentry{max confidence softmax}
\addplot [semithick, color2]
table {%
0 60.8027801513672
1 66.173942565918
2 64.8399200439453
3 67.6939544677734
4 69.6121826171875
5 70.5611343383789
6 69.6447982788086
7 71.8100433349609
8 71.4807815551758
9 72.1417694091797
10 72.1150054931641
11 72.3566818237305
12 72.8061370849609
13 72.8737106323242
14 72.7296142578125
15 73.0618057250977
16 72.7955627441406
17 73.8466796875
18 72.845947265625
19 72.6147079467773
20 72.8959808349609
21 73.6632080078125
22 73.1587829589844
23 73.4833297729492
24 72.8645858764648
25 72.9840393066406
26 73.3526611328125
27 73.5308990478516
28 73.4861907958984
29 73.3843612670898
30 73.4552536010742
31 73.4396209716797
32 73.4121398925781
33 73.59814453125
};
\addlegendentry{soft voting}
\end{axis}

\end{tikzpicture}

%% file: 4_related_work.tex
\section{Related Work}
\label{sec:related_work}

This section briefly describes other work in the areas of semantic segmentation and semi-supervised learning.

\paragraph{Semantic segmentation.}
Semantic segmentation is a fundamental problem in computer vision, which consider assigning labels to each pixel in an image.
Modern deep neural were networks successfully adapted to this problem, with fully-convolutional networks (FCN) \cite{long2015fully} being one of the first and most influential approaches.
Many solutions follow the encoder-decoder architecture, such as U-Net \cite{ronneberger2015u}.
Numerous techniques have been developed, such as pyramid scene parsing in PSPNet \cite{zhao2017pyramid}, or dilated convolutions and atrous spatial pyramid pooling as in DeepLabv3+ \cite{deeplabv3plus2018}.
Another architecture, HRnet \cite{wang2020deep} is focused on keeping high-resolution representations during the training.

\paragraph{Semi-supervised learning.} 
A recent line of research has shown that deep neural networks can be successfully used for semantic segmentation, given the right amount of data.
From the practical point of view, this is a challenging problem due to the high cost of labelling.
The goal of semi-supervised learning is to learn a model on a dataset, for which the labels are known on a certain percentage of the data.
Semi-supervised learning combines semantic segmentation and semi-supervised learning.
Recently, an intense effort can be observed in developing two families of techniques: contrastive learning and consistency regularisation.
Contrastive learning is built around learning representations which are \emph{close} for the samples of the same class and \emph{far} otherwise.
Consistency regularisation assumes that the same samples should yield the same labels under different -- often heavy -- augmentations and perturbations.
The disagreement between models is later used for the training.
 
Apart from CPS \cite{chen2021semi}, there are several architectures dedicated to semi-supervised learning.
A somewhat similar apporaches are cross-consistency training (CCT) \cite{ouali2020semi} or GCT \cite{ke2020guided}, which uses cross-confidence consistency for feature perturbation.
Mean Teacher \cite{tarvainen2017mean} considers a setting in which two models (student and teacher) are trained on the same dataset but using different augmentations.
The student model is trained in a standard way, while the teacher model is an exponential moving average of student models from previous steps.
The latter is also responsible for generating pseudo labels.
The Mean Teacher framework combined with CutMix \cite{yun2019cutmix} was used for semi-supervised segmentation in CutMix-Seg \cite{french2019semi}.
In Dynamic Mutual Training (abbreviated as DMT) \cite{feng2020dmt}, two neural networks are trained using a dynamically re-weighted loss function.
The authors of DMT leverages the disagreement between the models, which indicates a possible error and lowers the loss value.
ReCo (an abbreviation from Regional Contrast) \cite{liu2021bootstrapping} is a pixel-level contrastive learning framework, which incorporates memory-efficient sampling strategies.
The framework proved to be very effective in few-supervision scenarios, reaching 50\% mIoU on Cityscapes while requiring only 20 labelled images.

%% file: 5_summary.tex
\section{Summary}
\label{sec:summary}

In this paper, we presented $n$-CPS -- a generalisation of the consistency regularisation framework CPS.
We also proposed to utilise all the learned subnetworks for evaluation purposes using the ensemble learning techniques.
Evaluation of our approach on the Pascal VOC dataset showed that it sets the new state-of-the-art in its category.
An unavoidable limitation of such a study stems from the large number of parameters involved in the models.
In practical settings, this means relatively large GPU memory requirements.
Further work should consider the evaluation of ResNet-101 on the Cityscapes dataset, as the results on ResNet-50 are promising.
It should also consider evaluating the behaviour of models with $n \geq 4$ for both ResNet-50 and ResNet-101.
Finally, higher values of $n$ might also be tested using smaller networks (such as ResNet-18) to explore the behaviour of the models and the influence of ensembling techniques in such conditions.

%% file: 6_acknowledgements.tex
\section*{Acknowledgements}
Some experiments were performed using the Entropy cluster at the Institute of Informatics, University of Warsaw, funded by NVIDIA, Intel, the Polish National Science Center grant UMO2017/26/E/ST6/00622 and ERC Starting Grant TOTAL.

\section*{CRediT author statement}
\textbf{Dominik Filipiak} (80\% of the work): Conceptualisation, Methodology, Software, Validation, Formal Analysis, Investigation, Resources, Writing -- Original Draft, Writing -- Review \& Editing, Visualization, Project Administration.
\textbf{Piotr Tempczyk} (10\% of the work): Conceptualisation, Funding Acquisition, Supervision, Writing -- Review \& Editing.
\textbf{Marek Cygan} (10\% of the work): Conceptualisation, Supervision, Resources, Writing -- Review \& Editing. 